\documentclass[conference]{IEEEtran}
\IEEEoverridecommandlockouts
\usepackage{cite}
\usepackage{amsmath,amssymb,amsfonts}
\usepackage{algorithmic}
\usepackage{graphicx}
\usepackage{textcomp}
\usepackage{xcolor}
\def\BibTeX{{\rm B\kern-.05em{\sc i\kern-.025em b}\kern-.08em
    T\kern-.1667em\lower.7ex\hbox{E}\kern-.125emX}}

\usepackage{subcaption}
\usepackage{multirow}

\newcommand{\sign}{\mbox{sign}}
\newcommand{\relu}{\mbox{ReLU}}
\newcommand{\Var}{\mathrm{Var}}
\begin{document}

\title{A non-discriminatory approach\\to ethical deep learning
}

\author{\IEEEauthorblockN{Enzo Tartaglione}
\IEEEauthorblockA{\textit{Computer Science dept.} \\
\textit{University of Torino}\\
Torino, Italy \\
enzo.tartaglione@unito.it}
\and
\IEEEauthorblockN{Marco Grangetto}
\IEEEauthorblockA{\textit{Computer Science dept.} \\
\textit{University of Torino}\\
Torino, Italy}
}

\maketitle

\begin{abstract}
Artificial neural networks perform state-of-the-art in an ever-growing number of tasks, nowadays they are used to solve an incredibly large variety of tasks. However, typical training strategies do not take into account lawful, ethical and discriminatory potential issues the trained ANN models could incur in.\\
In this work we propose NDR, a non-discriminatory regularization strategy to prevent the ANN model to solve the target task using some discriminatory features like, for example, the ethnicity in an image classification task for human faces. In particular, a part of the ANN model is trained to hide the discriminatory information such that the rest of the network focuses in learning the given learning task. Our experiments show that NDR can be exploited to achieve non-discriminatory models with both minimal computational overhead and performance loss.
\end{abstract}


\section{Introduction}
\label{sec:introduction}
In the last two decades artificial neural network models (ANNs) received huge interest from the research community. Their capability of solving extremely complex tasks with simple heuristic approaches made them a potentially ``universal problem solving tool'', particularly in the area of supervised learning problem. Nowadays, complex and even ill-posed problems can be tackled provided that one can train a deep enough ANN model with a large enough dataset. Furthermore, they aim to become a powerful tool helping us taking decisions: for example, AI is currently used for scouting and hiring people~\cite{laumer2015impact}. These ANNs are trained to process a desired output from some inputs. We have no idea how the information is effectively processed inside. 
Recently, AI trustworthiness has been recognized as major prerequisite for people and societies to use and accept such systems ~\cite{zhang2019artificial,hleg2019ethics}. In April 2019, the High-Level Expert Group on AI of the European Commission defined the three main aspects of trustworthy AI \cite{hleg2019ethics}: it should be lawful, ethical and robust.
In particular, \cite{hleg2019ethics} presents, among the fundamental rights at the basis for Trustworthy AI,  ``equality, non-discrimination and solidarity, including the rights of persons at risk of exclusion [...]. In an AI context, equality entails that the system's operations cannot generate unfairly biased outputs (e.g. the data used to train AI systems should be as inclusive as possible, representing different population groups)''.
How to guarantee from a technical point of view that an ANN model is respecting those ethical guidelines remains an open question. In this work we focus on plugging ethical constraints in the learning process, e.g. to avoid  
gender and race biases that have been found in commercial face recognition tools~\cite{buolamwini2018gender, corbett2018measure} or leakage of private information~\cite{mo2019differential}. As another example, \cite{zhang2019artificial} 
suggests that ``preventing AI-assisted surveillance from violating privacy and civil liberties'' is an absolute priority.

In this work we are introducing a regularization term whose goal is to enforce an ethical constraint by hiding discriminatory features like race, gender, etc.
In other words, we aim at guaranteeing that discriminatory information is not exploited to infer the output of the model. 
\begin{figure}
    \centering
    \includegraphics[width=0.6\columnwidth]{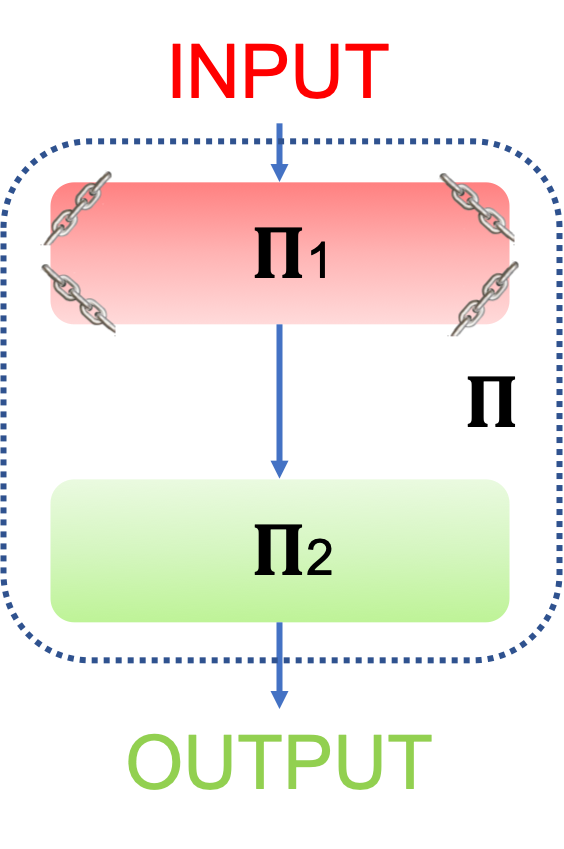}
    \caption{General scheme of an ANN $\Pi$: we can divide it in two sub-networks $\Pi_1$ and $\Pi_2$. While $\Pi_1$ contains discriminatory information (it learns how to clean input data from discriminatory features), $\Pi_2$ takes $\Pi_1$'s discriminatory features-free output and processes it to learn the training task.}
    \label{fig:genscheme}
\end{figure}
Towards this end let us consider an ANN model $\Pi$ as the concatenation of two sub-networks, $\Pi_1$ and $\Pi_2$ (as in Fig.~\ref{fig:genscheme}):
\begin{itemize}
    \item $\Pi_1$ will process the input to provide, as output, \emph{non discriminatory} 
    feature vectors.
    \item $\Pi_2$ will compute the final output using the only the non discriminatory features from $\Pi_1$.
\end{itemize}
According to the proposed approach the discriminatory information is washed-out by the processing in $\Pi_1$  
thus guaranteeing that training of $\Pi_2$ does not depend on discriminatory features of the input. Moreover, the same partitioning can be exploited when the pre-trained model is deployed in a system or final application: in a such a case one knows what part of the ANN cannot be compromised for ethical reasons. As an example $\Pi_1$ can be computed on board a secured device, e.g. personal mobile or edge computing terminal, and only non discriminatory features are sent to $\Pi_2$ implemented by a remote cloud service.
Our major contribution in this work is the proposal of a new non-discriminatory regularization (NDR) term to be enforced during the training of the model. The goal is to train $\Pi_1$ to hide some discriminatory features such that any task performed by $\Pi_2$ will not be discriminatory for those features. The proposed NDR is shown to be computationally-efficient: it introduces a negligible computational overhead during training, proportional to the cardinality of the ``discriminatory classes''. Intuitively, as NDR hides some potentially-useful information, the learning task in $\Pi_2$ might become harder to solve: this is the price to be paid to achieve AI trustworthiness, which might result into an acceptable compromise in term of performance drop. Such a problem can not be tackled directly attempting to hide the mutual information between samples in the same discriminatory class: the non-differentiability of such a measure as well as the introduced computational complexity is a huge obstacle, and NDR proposes itself as a proxy for such a measure. Nonetheless, previous works have already shown that adding further constraints to the learning problem could be effective~\cite{tartaglione2019post} as, typically, the trained ANN models are over-sized and allows a large number of solutions to the same learning task~\cite{tartaglione2019take}. Our experiments with NDR show that in practical cases it is possible to strike a good balance between non discriminatory constraint and target performance.\\
The rest of the work is structured as follows. In Sec.~\ref{sec:related} we review some works close to our problem. Then, in Sec.~\ref{sec:method} we present our NDR term, giving a justification and analyzing in depth the expected effect on the ANN model (and in particular, to $\Pi_1$). Then, in Sec.~\ref{sec::results} some empirical results are shown and finally, in Sec.~\ref{sec:conclusion}, the conclusions are drawn.
\section{Related works}
\label{sec:related}

In this section an overview of some closely related-to non-discriminatory learning topics will be presented. Because of the specific nature of our aim, there is still no available literature solving exactly the non-discriminatory learning problem, even though the problem has already been mentioned from an information-leakage perspective~\cite{shmatikovmachine}. However, there are some topics which are closely related to it: those include, but are not limited to:
\begin{itemize}
    \item bias issues: some outputs can be affected by internal biases the dataset intrinsically incorporates;
    \item fairness during learning: there might be some unbalancing between classes, and the class having the highest population dominates;
    \item ensuring privacy.
\end{itemize}
It is known that datasets are typically affected by biases. In their work, Torralba~and~Efros~\cite{torralba2011unbiased} showed some biases affecting some of the most used datasets, drawing considerations on generalization performance and classification capability of the trained ANN models. Following the same approach, Tommasi~\emph{et al.}~\cite{tommasi2017deeper} conducted a series of experiments reporting differences between a number of datasets and verifying differences in final performance when applying different de-biasing strategies, aiming at balancing data. Working at the dataset level is critical in general, and helps a lot in understanding the data and its criticalities~\cite{Cubuk_2019_CVPR}. The concept of removing bias by using data borrowed by different sources has been explored in a practical and empirical context by Gupta~\emph{et~al.}~\cite{gupta2018robot}. In particular, they designed an un-biasing strategy to minimize the effects of imperfect execution and calibration errors by balancing the effect of unbalanced data, showing improvements in the generalization of the final model.\\
Dataset un-biasing helps in the learning process, as training is performed with no biases; however, with such an approach typically we have no control on the information we are removing from the dataset itself, and we are not guaranteeing the final architecture will not be driven by a certain subset of biases (like, for example, races or sex differences). A context in which, on the contrary, we can have direct access to these biases is presented by Hendricks~\emph{et~al.}~\cite{hendricks2018women}. In such a work it was possible to explicitly introduce a corrective loss term (coherent with the formulation introduced by Vinyals~\emph{et al.}~\cite{vinyals2015show}) with the aim to help the ANN model to focus on the correct features.\\
Having an explicit formulation for the loss term is typically not possible. To overcome this problem, a number of strategies have been proposed, among which generative adversarial networks-based solutions cover a prominent role. Some works suggest the use of GANs to entirely clean-up the dataset with the aim of providing fairness~\cite{xu2018fairgan, sattigeri2018fairness}, others like Mandras~\emph{et~al.}~\cite{madras2018learning} insert a GAN in the middle of the architecture to clean-up the internal representation of data. In general, training such an architecture is a very delicate and complex process, and it does not provide explicitly fairness in the inference phase, as generative adversarial networks are used to generate training data. Besides GAN-based approaches, we need to mention the work by Beutel~\emph{et~at.}~\cite{beutel2019putting}, where conditional equality has been introduced as fairness metrics during the training phase, aiming at improving the final performance of the trained model. Differently, Orekondy~\emph{et~al.}~\cite{Orekondy_2018_CVPR} proposed an approach to clean-up datasets, removing some ``private'' information but still maintaining the utility of the dataset for the trained task.\\
A recent work by Song~\emph{et~al.}~\cite{song2017machine} reported that, using standard training strategies to train some state-of-the-art models, allows information not relevant to the learning task to be stored inside the network. Such a behavior is possible because of the typically oversized ANNs trained to solve a task~\cite{tartaglione2019take}. In their experiments, Song~\emph{et~al.} show how accurately they can recover some non-directly related to training information. In this way, they aim to show the potential lack of privacy in these models. Schmatikov and Song~\cite{shmatikovmachine}, furthermore, showed how easily we can impose data, during the learning process, to be sub-clustered, and how easily we can recover secondary features not-directly related to the learning itself. The theme of privacy is not novel in the field~\cite{hassan2019automatic}: approaches to sophisticated infrastructures have indeed been proposed~\cite{das2017assisting} or variants to SGD which are privacy-aware~\cite{Wu_2019_CVPR}, or yet GAN-based solutions~\cite{Roy_2019_CVPR}. A different approach to solve privacy issues has been proposed by Chabanne~\emph{et~al.}, where the authors revise Cryptonets~\cite{gilad2016cryptonets} (they essentially crypt the input information and perform operations on top of those) and make them scalable with the use of batch-normalization.\\
In the following section we introduce non-discriminatory regularization (NDR), a novel regularization strategy to be paired with the state-of-the-art deep learning training strategies, whose aim is to prevent the ANN model to take advantage of some \emph{discriminatory features} to perform the given learning task. Such an aim clearly differs from fairness and bias issues because we are hiding part of the information while in the literature those issues are solved attempting to balance the information. Regarding privacy, our topic is clearly different: while we focus on the output not being affected by some features, privacy issues refer to the input not to be back-traceable.

\section{Non-Discriminatory Regularization}
\label{sec:method}
In this section, after introducing the notation, we present NDR, a new regularization term, whose aim is to hide some \emph{discriminatory features} for a given task. In particular, NDR will be applied at the output of a given sub-network $\Pi_1$, which will be the input for a sub-network $\Pi_2$ computing the output of the ANN model, as already described in Sec.~\ref{sec:introduction}. NDR is designed to promote uncorrelated feature vectors belonging to the same discriminatory class: in such a way, the discriminatory information will be hidden during training. Hence, during the training phase, the objective function we aim to minimize will be
\begin{equation}
    \label{eq:uprule}
    J = \eta L + \gamma R_l
\end{equation}
where $L$ is the loss function for the trained task and $R_l$ is the proposed NDR term applied at $l$-th output layer, i.e. the $\Pi_1$ output.  Clearly, $\eta$ and $\gamma$ are two positive hyper-parameters. 
Finally. we are going to investigate the actual contribution of $R_l$ during back-propagation in $\Pi_1$.

\subsection{Preliminaries}
\begin{figure}
    \centering
    \includegraphics[width=0.8\columnwidth]{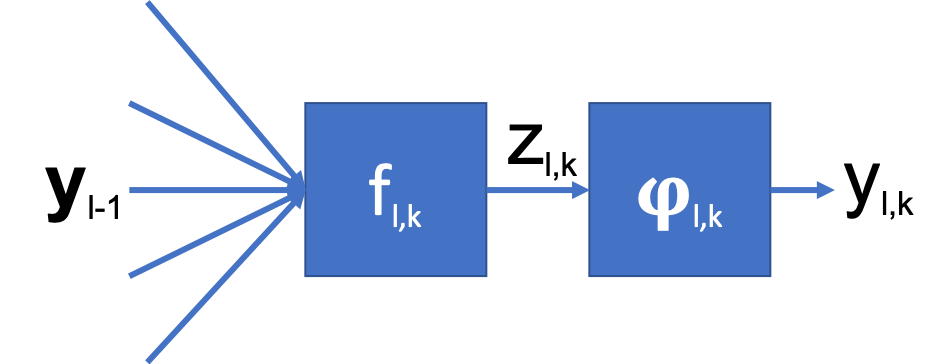}
    \caption{Visual representation of the notation used at the neuron level.}
    \label{fig:notation}
\end{figure}
In this section we introduce the notation we are going to use for the rest of this work. Let us assume we work with an acyclic, multi-layer artificial neural network composed of $L$ layers, where layer $l=0$ is the input layer and $l=L$ the output layer. Each of the $l$ layers is made of $K_l$ neurons (or filters for convolutional layers). Hence, the $k$-th neuron ($k\in [1, K_l]$) in the $l$-th layer has:
\begin{itemize}
    \item $y_{l, k}$ as its own output.
    \item $\mathbf{y}_{l-1}$ as input vector.
    \item $\mathbf{w}_{l, k}$ as the weights (from which we identify the $p$-th as $w_{l, k, p}$) and $b_{l, k}$ as the bias.
\end{itemize}
Each of the neurons has its own activation function $\varphi_{l, k}(\cdot )$ to be applied after some affine function $f_{l, k}(\cdot)$ which can be, for example, a convolution or a dot product. Hence, the output of a neuron can be expressed by
\begin{equation}
    \label{eq::defy}
    y_{l, k} = \varphi_{l, k}\left[ f_{l, k} \left(\mathbf{w}_{l, k}, b_{l, k}, \mathbf{y}_{l-1} \right) \right]
\end{equation}
Each layer, during the forward-propagation step, will output a tensor $\mathbf{y}_l \in \mathbb{R}^{N_l \times M}$, where
\begin{itemize}
    \item $M$ is the mini-batch size.
    \item $N_l$ is the output size of the layer for a single input pattern.
\end{itemize}

\subsection{Discriminatory data correlations}
\label{sec:ddc}
Let us consider the $l$-th layer in the ANN model, representing the output layer of $\Pi_1$. Clearly, $\mathbf{y}_l$ is a representation of the input samples. If we assume these data can be grouped in $C$ \emph{discriminatory classes}, our mission is to train $\Pi_1$ such that those classes are not recognizable at the output of the $l$-th layer. Towards this end, we are proposing a two stages approach:
\begin{itemize}
    \item normalize each feature vector $\mathbf{y}_l$: in such a way, the norm will be the same for all the  representation;
    \item make all the feature vectors, belonging to the same discriminatory class, orthogonal.
\end{itemize}


We define $\mathbf{y}_l^c$ as the exclusive subset of the outputs at layer $l$ which belong to the same dicriminatory class $c$.
We can build a \emph{similarity matrix} $G_{l}^c \in \mathbb{R}^{M^c \times M^c}$, where $M^c$ is the cardinality of the data belonging to the $c$-th discriminatory class: 
\begin{equation}
    G_{l}^c = \left(\tilde{\mathbf{y}}_{l}^c\right)^T \cdot \tilde{\mathbf{y}}_{l}^c
\end{equation}
where $(\cdot)^T$ indicates transposed matrix and $ \tilde{\mathbf{y}}_{l}^c$ indicates a per-representation normalization
\begin{equation}
    \tilde{\mathbf{y}}_{l,i}^c = \frac{\mathbf{y}_{l,i}^c}{\| \mathbf{y}_{l,i}^c \|_2} \forall i \in [1, M^c]
\end{equation}
Hence, every $g_{l, i, j}^c$ entry between two patterns $i,j$ in $G_{l}^c$ indicates their correlation:
\begin{equation}
    g_{l,i,j}^c = \left(\tilde{\mathbf{y}}_{l,i}^c \right)^T \cdot \tilde{\mathbf{y}}_{l,j}^c
\end{equation}
$G_{l}^c$ is a special case of \emph{Gramian matrix} as any $g_{l,i,j}^c \in [-1;+1]$ and indicate the difference in the direction between any two $\mathbf{y}_{l,i}^c$ and $\mathbf{y}_{l,j}^c$:
\begin{equation}
    \beta_{l,i,j}^c = \arccos{g_{l,i,j}^c}
\end{equation}
$G_{l}^c$ has some properties:
\begin{itemize}
    \item is a symmetric, positive-semidefinite matrix.
    \item all the elements in the main diagonal are exactly 1 by construction.
    \item if the subset of outputs $\tilde{\mathbf{y}}_{l}^c$ form an orthonormal basis (or, in this context, equivalently $G_{l}^c$ is full-rank), then the final matrix will be an identity by definition.
\end{itemize}
Thus, if $G_{l}^c$ is an identity matrix, then no common features for the $c$-th class can be identified, ensuring non-discrimination.

\subsection{Non-Discriminatory Regularization}
In Sec.~\ref{sec:ddc} it has been shown a way to find correlations between feature vectors. In order to obtain $G_l^c \rightarrow \mathbb{I}\ \forall c$, we can push all the off-diagonal elements to zero with the following regularization term:
\begin{equation}
    \label{eq::strongregu}
    R_{l}^{strong} = \frac{1}{C}\sum_{c} \frac{1}{2 M^c}\sum_{i\neq j} \left| g_{l,i,j}^c \right|
\end{equation}
Having $R_{l}^{strong} \rightarrow 0$ is a hard constraint. 
We can impose a weaker, still effective condition based on the average of the correlations:
\begin{equation}
    \label{eq::regucond}
    R_{l} = \frac{1}{C}\sum_c \frac{1}{2 M^c}\left| \sum_{i\neq j} g_{l,i,j}^c \right|
\end{equation}
Notice that for ReLU-activated networks the weaker and the strong condition coincide, as $g_{l,i,j}^c \geq 0\ \forall i,j$. 

\subsection{Back-propagating NDR}
In the following we analize the effect of the regularization imposed in \eqref{eq::regucond}.
For sake of simplicity, we focus on the contribution of the inner summation in \eqref{eq::regucond} related to single class $c$: $R^c_l=\frac{1}{2 M^c}\left| \sum_{i\neq j} g_{l,i,j}^c \right|$.
Using the chain rule to compute the derivative over the $q$-th weight $w_{l,n,q}$ (belonging to the $n$-th neuron at the $l$-th layer), we get:
\begin{equation}
    \frac{\partial R_{l}^c}{\partial w_{l,n,q}} = -\frac{1}{2 M^c V_{l,i}^c V_{l,j}^c}\sign(R_{l}^c) 
    \frac{{y}_{l,i,n}^c}{w_{l,n,q}} \frac{{y}_{l,j,n}^c}{w_{l,n,q}}
\end{equation}
where
\begin{equation}
    V_{l,i}^c = \|\mathbf{y}_{l,i}^c\|_2
\end{equation}
If we define here
\begin{equation}
    \Delta_{i}^c = \frac{1}{M^c \cdot V_i^c} \varphi_{l,n}' [z_{l,i,n}^c] y_{l-1,i,q}^c
\end{equation}
we find that
\begin{equation}
    \label{eq:weightupdate}
    \frac{\partial R_{l}^c}{\partial w_{l,n,q}} = -\frac{1}{2}\sign(R_{l}^c) \Var\left[\mathbf{\Delta}^c \right]
\end{equation}
where $\Var[ \mathbf{\Delta}^c ]$ indicates the variance of all the $\Delta_{i}^c$ values. 
Now let us analyze the minimum we are targeting: the derivative \eqref{eq:weightupdate} is equal to zero only when $\Delta_{i}^c = \Delta_{j}^c\ \forall i, j$. Assuming $\varphi(\cdot) = \relu(\cdot)$ (which is the most common case in state-of-the-art architectures) we can discuss the following cases:
\begin{itemize}
    \item case $\Delta_{i}^c \neq 0$:  we need $\varphi_{l,n}' [z_{l,j,n}^c] = 1$ and, necessarily $\frac{y_{l-1, i, q}^c}{V_{l,i}^c} = \frac{y_{l-1, j, q}^c}{V_{l,j}^c} \forall i, j$. However, this is an extremely-unlikely case, as it implies the $n$-th neuron being in the linear region for all the examples, and implies a very specific condition we just need to avoid during training.
    \item case $\Delta_{i}^c = 0$: we can achieve zero derivative guaranteeing one of the three following conditions:
    \begin{itemize}
        \item $V_{i}^c \rightarrow + \infty$, which however should be avoided, e.g. by enforcing a weight decay regularization;
        \item $\varphi_{l,n}' [z_{l,j,n}^c] = 0$, meaning that the neuron is in the non-linear region, hence turned off;
        \item if none of the above conditions is satisfied, the input signal to the $l$-th layer $y_{l-1, i, q}^c$ is changed through back-propagation. 
    \end{itemize}
\end{itemize}
The proposed analysis shows how NDR is impacting on the model weights, promoting orthogonalization of the representations of a single discriminatory class.  Of course, the same gradient contribution comes from all the $C$ discriminatory classes we aim to hide. In the next section we are going to show experimental results in practical cases where NDR is imposed with standard mini-batch approach.

\section{Results}
\label{sec::results}
In this section we experiment with our proposed method over some different neural architectures and datasets.
While training the ANN model on a target classification task, we identify some \emph{discriminatory} information we wish to hide (like, for example, the ethnicity in Sec.~\ref{sec:UTK}). Our training and inference algorithms are implemented in Python, using PyTorch~1.3 and a RTX2080 Ti NVIDIA GPU with 11GB of memory has been used for training and inference.\footnote{The source code will be made available upon acceptance of the article.} All the used hyper-parameters have been tuned using a grid-search algorithm.\\
The results will be presented for two different datasets: MNIST and UTKFace. Besides the final generalization performance of the trained model on a target task (for example, recognizing the classes odd/even for MNIST), an empirical analysis on the level of correlation remaining in the output of the regularized layer has been performed. In particular, the aim is here to retrieve the discriminatory information. In order to do this, following a similar approach as the one proposed by Schmatikov~and~Song~\cite{shmatikovmachine}, the output of the layer where NDR is applied is processed as follows:
\begin{itemize}
    \item because of the typical high-dimensionality of the output for a standard ANN architecture layer (from hundreds to millions of dimensions), PCA is applied to reduce the dimensionality of data. PCA implementation from the scikit-learn library has been used.
    \item then, K-means clustering is applied on the dimensionality-reduced data. In this case, as we already know the classes we are trying to hide, we a-priori know the optimal K to be chosen; for this reason, no study on the optimal K will be performed. Then, we look at the distance between the centroids computed with K-means and the \emph{discriminatory centroids}, i.e. the centroids computed on the discriminatory classes. Also the K-means algorithm used for clustering belongs to the scikit-learn library.
\end{itemize}

\subsection{Odd-even classification on MNIST}
\label{sec:MNIST}
\begin{figure*}[tb]
    \captionsetup[subfigure]{position=b}
    \subcaptionbox{~ \label{fig:CE_conv1}}{\includegraphics[width=.4\linewidth]{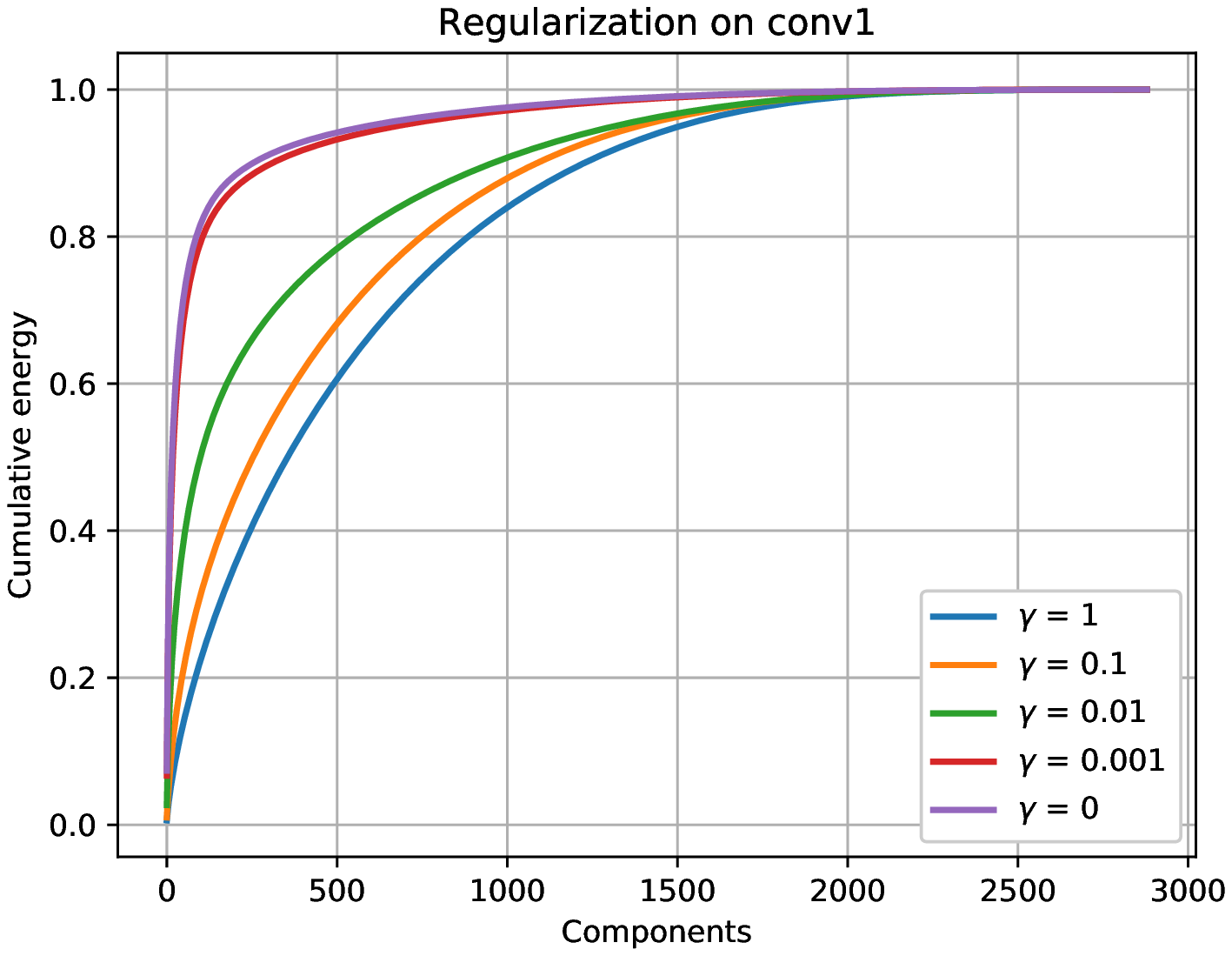}}
    \subcaptionbox{~ \label{fig:KM_conv1_g0}}{\includegraphics[width=.3\linewidth]{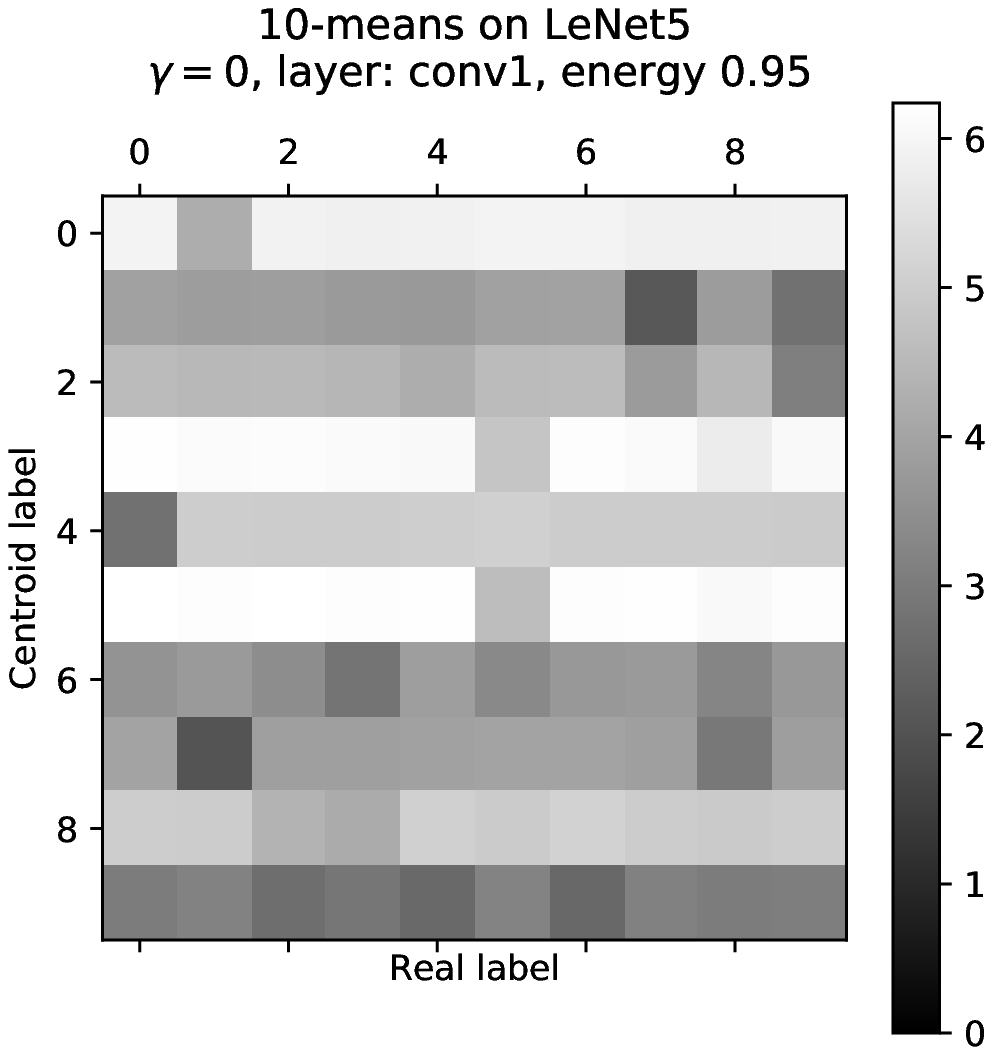}}
    \subcaptionbox{~ \label{fig:KM_conv1_g1}}{\includegraphics[width=.3\linewidth]{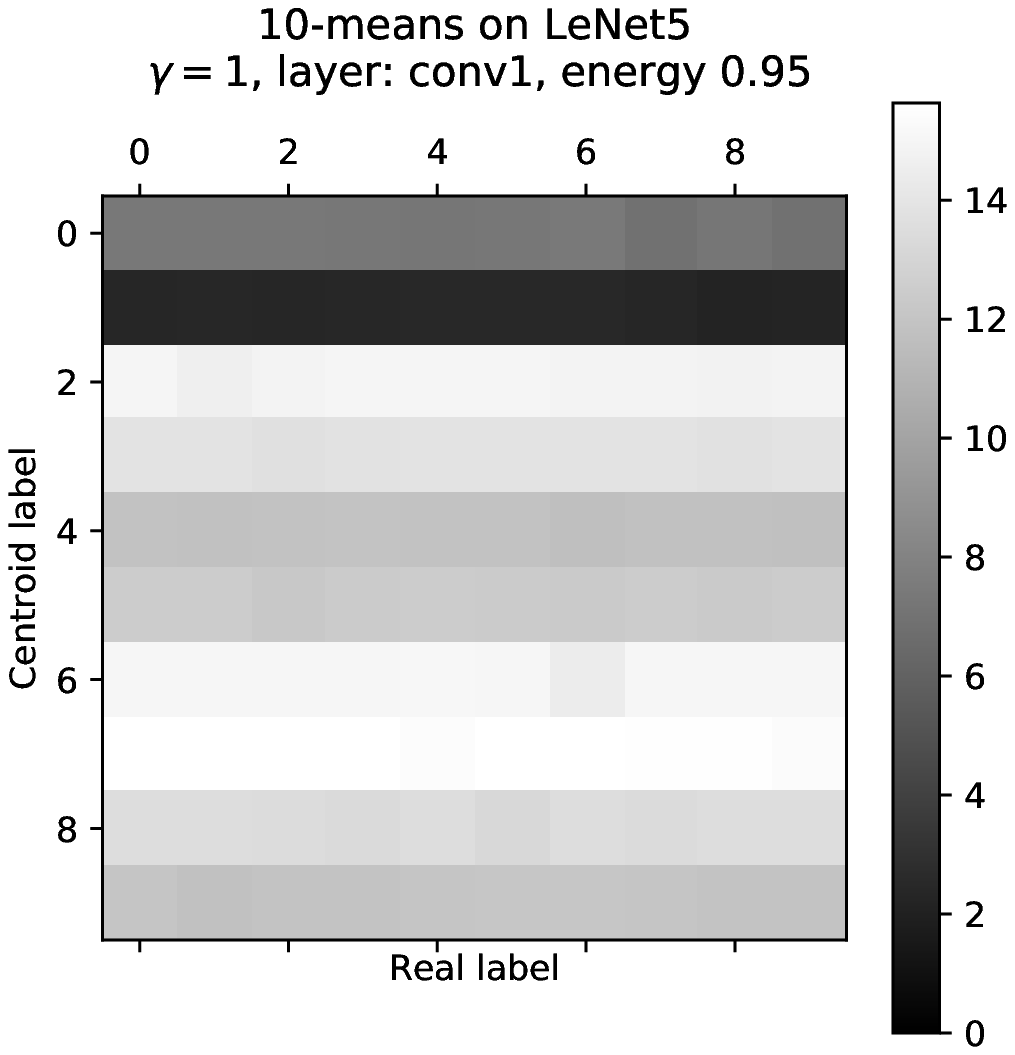}}
    \caption{Cumulative energy of the PCA eigenvalue components when PCA is applied on \texttt{conv1} output for LeNet5-caffe trained on MNIST (a), distance-matrix between K-means centroids and discriminatory centroids, $l$ is \texttt{conv1}, for $\gamma=0$ (b) and $\gamma=1$ (c), using a number of dimensions determined by the 95\% of the PCA eigenvalues energy.}
    \label{fig:KM_conv1}
\end{figure*}

\begin{figure*}[tb]
    \captionsetup[subfigure]{position=b}
    \subcaptionbox{~ \label{fig:CE_fc1}}{\includegraphics[width=.4\linewidth]{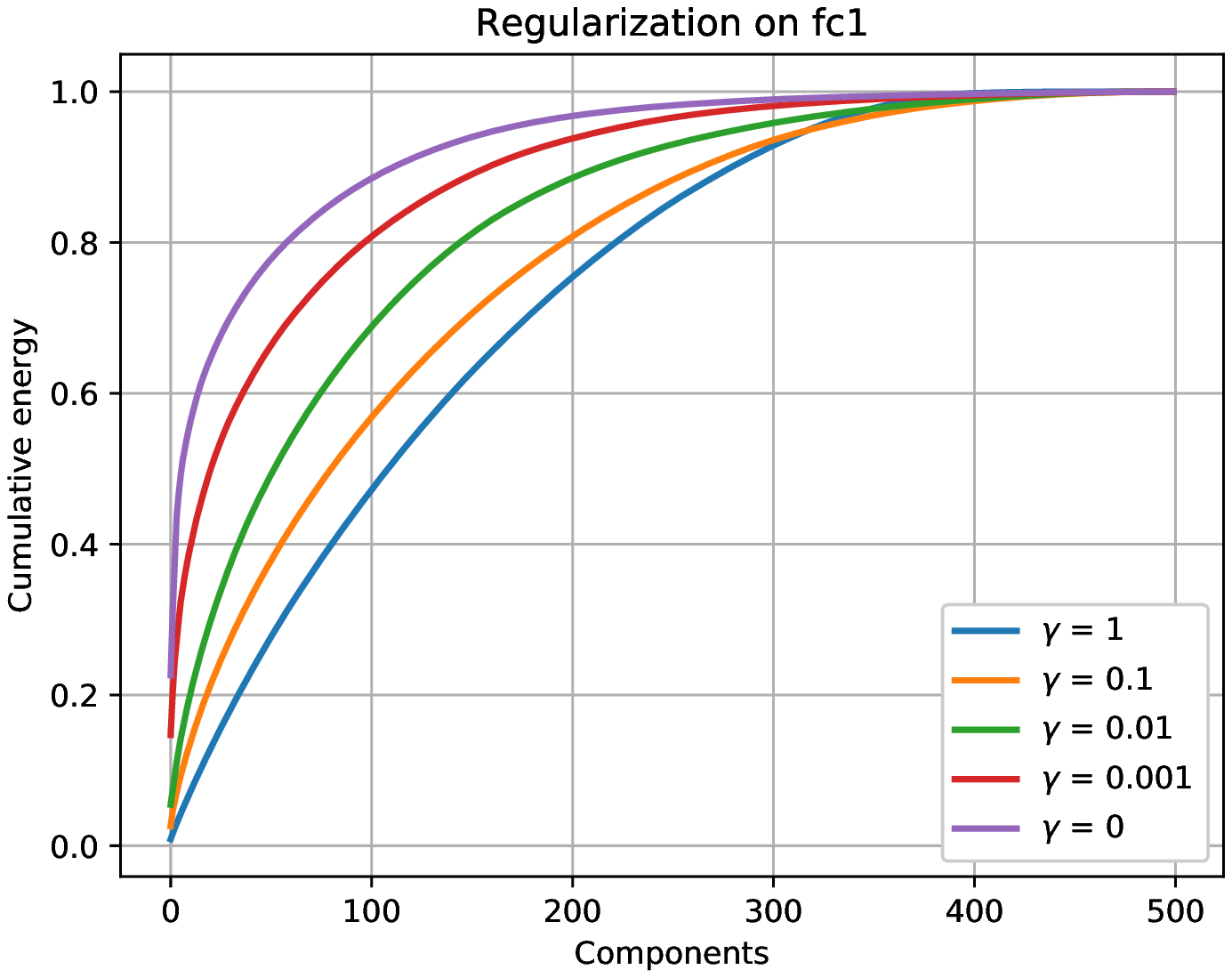}}
    \subcaptionbox{~ \label{fig:KM_fc1_g0}}{\includegraphics[width=.3\linewidth]{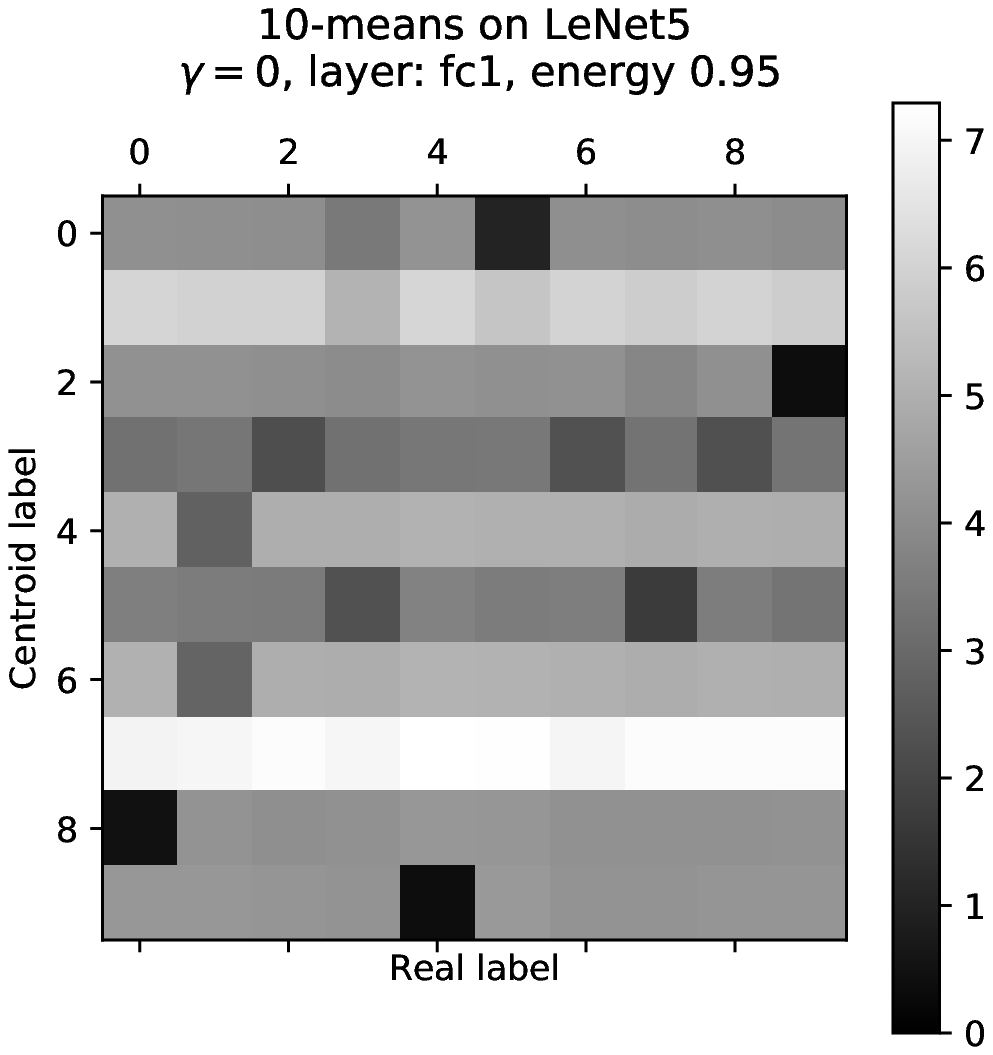}}
    \subcaptionbox{~ \label{fig:KM_fc1_g1}}{\includegraphics[width=.3\linewidth]{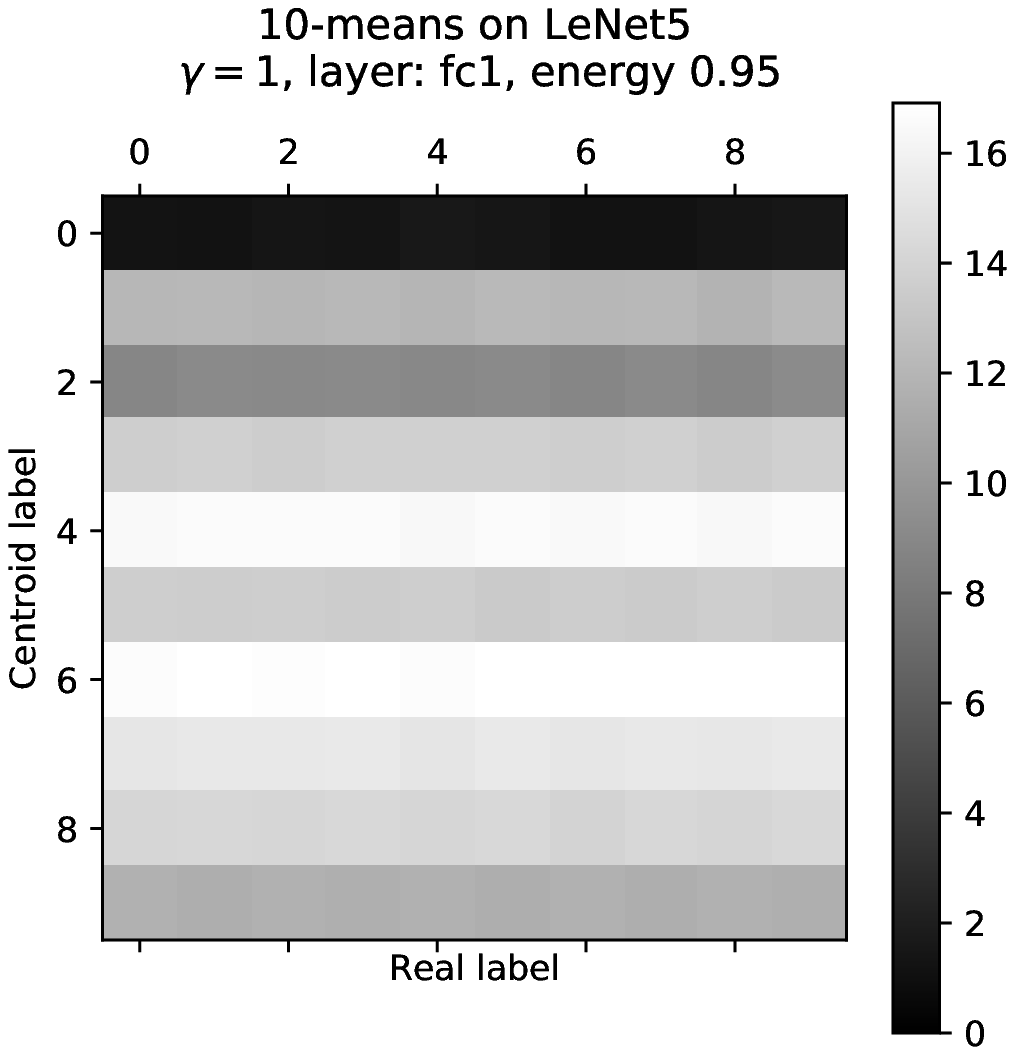}}
    \caption{Cumulative energy of the PCA eigenvalue components when PCA is applied on \texttt{fc1} output for LeNet5-caffe trained on MNIST (a), distance-matrix between K-means centroids and discriminatory centroids, $l$ is \texttt{fc1}, for $\gamma=0$ (b) and $\gamma=1$ (c), using a number of dimensions determined by the 95\% of the PCA eigenvalues energy.}
    \label{fig:KM_fc1}
\end{figure*}

\begin{table*}[!t]
    \captionsetup{justification=centering, labelsep=newline}
    \renewcommand{\arraystretch}{1.3}
    \caption{Performance classification and NDR on MNIST and UTKFace.}
    \label{tab:results}
    \centering
        \begin{tabular}{|c|c|c|c|c|c|c|}
            \hline
            Dataset & Architecture & $\Pi_1$ & $\Pi_2$ & $\gamma$ & final $R_l$ & test accuracy [\%] \\
            \hline\hline
            \multirow{10}{*}{MNIST}& \multirow{10}{*}{LeNet5-caffe}&\multirow{5}{*}{\texttt{conv1}} & \multirow{5}{*}{\texttt{conv2,fc1,fc2}}&0&0.54   &99.32\\
            && &  &0.001      &0.50       &99.33\\
            && &  &0.01       &0.22       &98.94\\
            && &  &0.1        &0.07       &97.41\\
            && &  &1          &0.04       &95.54\\
            \cline{3-7} 
            &&\multirow{5}{*}{\texttt{conv1,conv2,fc1}} & \multirow{5}{*}{\texttt{fc2}}&0   &0.61&99.32\\
            && &  &0.001      &0.34       &99.36\\
            && &  &0.01       &0.19       &99.30\\
            && &  &0.1        &0.11       &99.15\\
            && &  &1          &0.03       &93.23\\
            \hline
            \multirow{2}{*}{UTKFace}& \multirow{2}{*}{Inception-v3}&\multirow{2}{*}{\texttt{[...]-conv2d\_4a}} & \multirow{2}{*}{\texttt{Incept.A-[...]}}&0&  0.78 &93.84\\
            &&&&1 & 0.01 &91.75\\
            \hline
        \end{tabular}
\end{table*}
In our first experiment we decided to apply the regularization strategy on a modified version of the LeNet-5 architecture, trained on MNIST. MNIST is a handwritten digits dataset made of 60k 28x28 pixel grey-scale images of digits, divided in 10 classes (from 0 to 9). For this task, we have decided to train the ANN model in recognizing whether a number is odd or even (target classes), attempting to hide the original $C=10$ classes each image belongs to, i.e. 0,1,2... (discriminatory classes). Towards this end, as the target is here different from the usual MNIST problem, the LeNet5-caffe architecture has a single output neuron (in place of the usual 10), whose output determines whether the input represents an odd or an even number. All the simulations on MNIST have been conducted minimizing a binary cross-entropy loss with vanilla-SGD, $\eta$ = 0.01, weight-decay 1e-5, minibatch size 100 for 500 epochs.\\
First, we decided to apply our regularization as close as possible to the input. For LeNet5-caffe, the dimensionality of the output of the first layer (\texttt{conv1}) is very large (2880).  In Fig.~\ref{fig:CE_conv1} we show the cumulative energy of the PCA eigenvalues at the output of $\Pi_1$ for different values of $\gamma$ in \eqref{eq:uprule}, $\gamma=0$ representing the baseline case when the proposed NDR is disabled. We can clearly see that, imposing a larger $\gamma$ value, the energy of the eigenvalues gets distributed on more dimensions making it harder to recognize the discriminatory classes.
Let us try to retrieve the information we aim to hide using K-means clustering on the output of \texttt{conv1}: for these experiments, the number of PCA dimensions to use is decided considering the 95\% of the overall energy. From Fig.~\ref{fig:KM_conv1} we see the distance matrix computed between the centroids from K-means and the real centroids of the discriminatory classes. While for the case without NDR (Fig.~\ref{fig:KM_conv1_g0}) we can see that some clusters are found (like, for example, the number ``0'' is found by centroid number 4, the ``1'' is found by the centroid 7 and the ``7'' by  centroid 1), when NDR is enabled (Fig.~\ref{fig:KM_conv1_g1}) the $C$ classes can no longer be identified. Fig.~\ref{fig:KM_conv1_g1} clearly shows that none of the 10 clusters formed by K-means has a centroid close to reals ones: in fact, looking at the horizontal stripes in the distance matrix, it can be noted that every computed centroid is approximately equidistant from all the centroids of the discriminatory classes.  
This, however, comes at a cost: in Tab.~\ref{tab:results} we observe that, imposing a larger $\gamma$ value results in a performance loss.\\
We have also tried to apply NDR on the \texttt{fc1} layer (closer to the ANN output): this has a lower dimensionality, and for instance it should apparently be an harder task for NDR. Also in this case, we are showing the cumulative energy of the PCA eigenvalues for different values of $\gamma$, in Fig.~\ref{fig:CE_fc1}. When compared to Fig.~\ref{fig:CE_conv1}, it can be noted that even for the baseline ($\gamma = 0$) we see that the energy for the eigenvalues is less concentrated: this is an effect determined by the reduced output size of the considered layer.\\
As above we have tried to retrieve discriminatory information using PCA and K-means: the results are shown in Fig.~\ref{fig:KM_fc1}. Here, the discriminatory information leakage when $\gamma=0$ (Fig.~\ref{fig:KM_fc1_g0}) is extremely evident: K-means retrieves exactly the discriminatory centroids for class ``0'', ``4'', ``5'' and ``9''. This can be explained by the lower dimensionality of layer \texttt{fc1}, which eases the retrieval process. Still, using the NDR approach discriminatory centroids are hidden, as shown in Fig.~\ref{fig:KM_fc1_g1}. As for the previous experiment, in Tab.~\ref{tab:results} we show that increasing $\gamma$ results in a performance drop which is, however, significant only for larger values of $\gamma$ compared to the previous case. In this latter case $\Pi_1$, being deeper and with a larger number of parameters, can be trained to be more effective in hiding discriminatory information without affecting the final classification accuracy at the same time. 

\subsection{Race-free gender classification on UTKFace}
\label{sec:UTK}
\begin{figure*}[tb]
    \captionsetup[subfigure]{position=b}
    \subcaptionbox{~ \label{fig:CE_conv2d}}{\includegraphics[width=.4\linewidth]{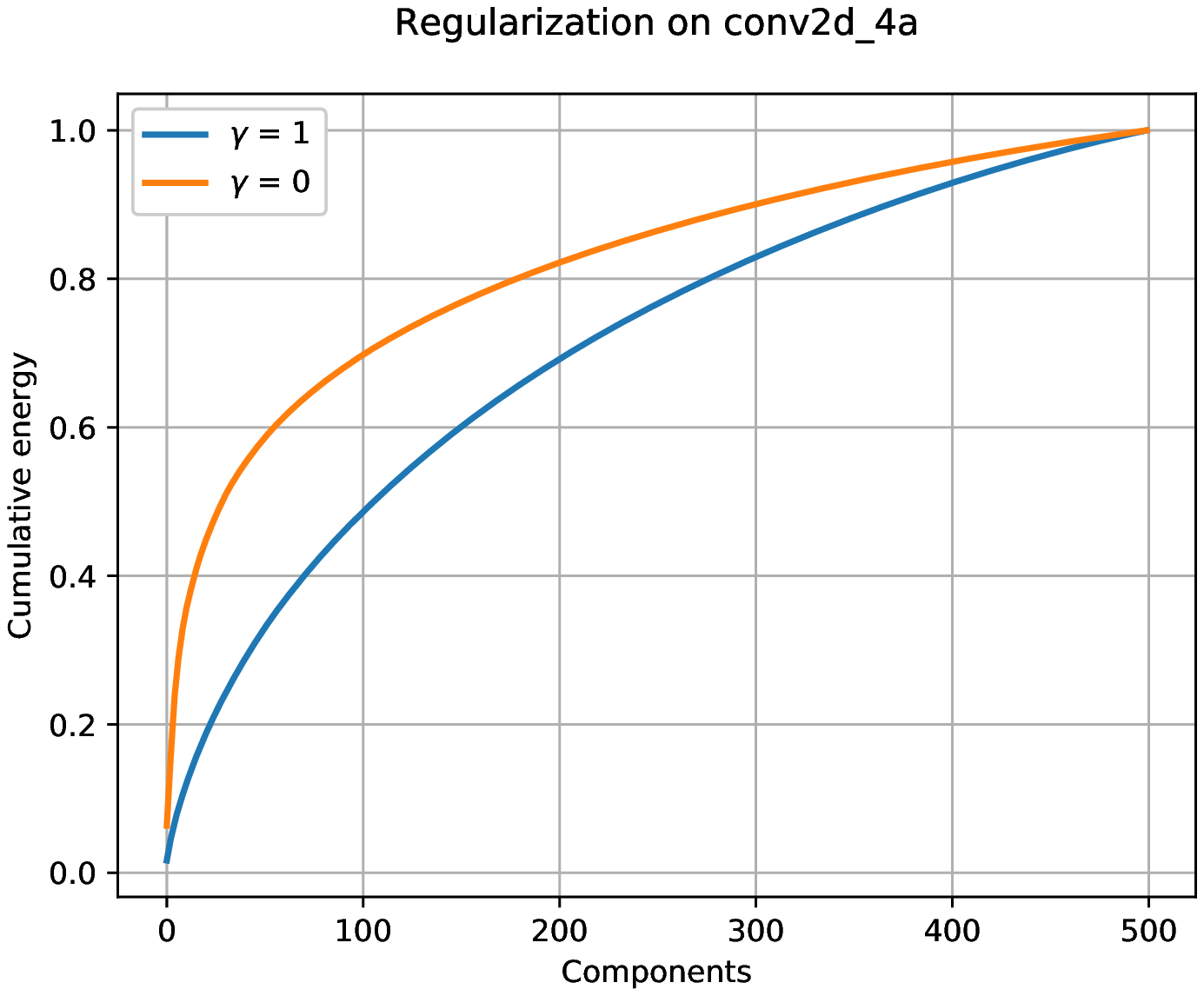}}
    \subcaptionbox{~ \label{fig:KM_conv2d_g0}}{\includegraphics[width=.3\linewidth]{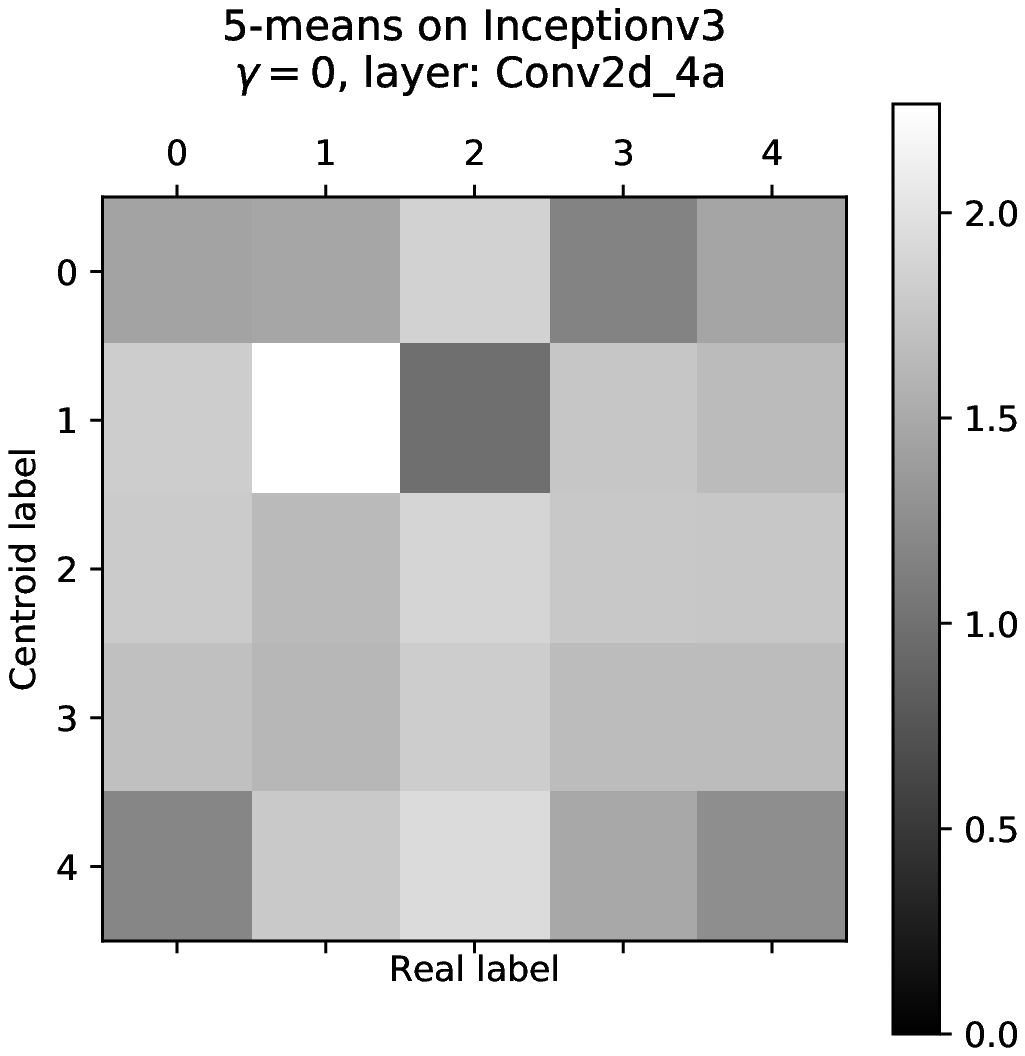}}
    \subcaptionbox{~ \label{fig:KM_conv2d_g1}}{\includegraphics[width=.3\linewidth]{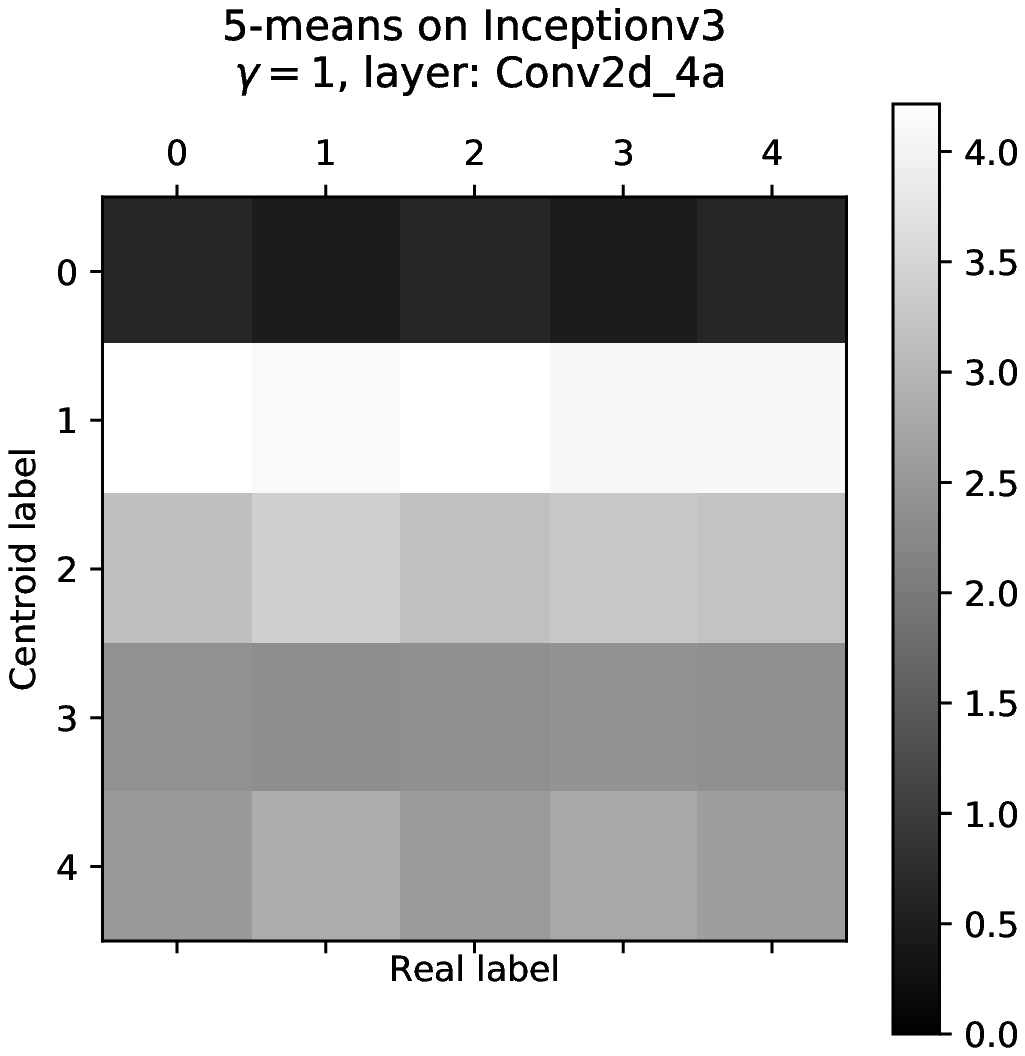}}
    \caption{Cumulative energy of the top 500 PCA eigenvalue components when PCA is applied on \texttt{conv2d\_4a} output for Inception~v3 trained on UTKFace(a), distance-matrix between K-means centroids and discriminatory centroids, $l$ is \texttt{conv2d\_4a}, for $\gamma=0$ (b) and $\gamma=1$ (c), after PCA with 10 components.}
	\label{fig:KM_conv2d}
\end{figure*}

We have also tried NDR in a more realistic discriminatory scenario, i.e. the UTKFace dataset, a large-scale face dataset. The dataset consists of over 20,000 200x200 RGB images of faces with annotations about age, gender, and ethnicity (5 ethnicities). 
In particular, we decided to focus on the ``non-discriminatory gender classification task'': the main task for the ANN model is to learn the face gender (2 target classes), without taking into consideration any information about ethnicity (5 discriminatory classes). The ANN we have chosen to use is Inception~v3~\cite{szegedy2016rethinking} 
that has been already proposed to solving similar tasks~\cite{schroff2015facenet}. All the simulations on UTKFace have been conduced minimizing a binary cross-entropy loss with vanilla-SGD, $\eta$ = 0.1, weight-decay 1e-4, minibatch size 50 for 100 epochs.\\
Given the complexity of the larger ANN model, we have decided to include in $\Pi_1$ the part of Inception~v3 before the first inception block, using NDR at the output of the \texttt{conv2d\_4a} layer. Because of the higher dimension of the output (968,000) we have just performed a partial PCA eigenvalues energy analysis, as shown in Fig.~\ref{fig:CE_conv2d}. It worth noticing that using NDR ($\gamma=1$) the energy is more evenly distributed than for the baseline ($\gamma=0$), matching what we observed for the simpler MNIST case. Moreover, when $\gamma=1$ the gender classification accuracy shown in Tab.~\ref{tab:results} remains acceptable with a limited reduction of about 2\%.
Also in this case we have run K-means to unveil the discriminatory clusters. Due to the high dimensionality of the problem, it was not possible to set a PCA eigenvalues energy boundary, as we did in Sec.~\ref{sec:MNIST}. 
For computational reasons, we have decided to reduce the dimensionality of \texttt{conv2d\_4a} to the first 10 components only, and then to apply K-means with $K=5$. The results are shown in Fig.~\ref{fig:KM_conv2d}. Similarly to the previous MNIST experiment, when NDR is applied K-means centroids does not match the discriminatory ones (Fig.~\ref{fig:KM_conv2d_g1}). On the contrary, for $\gamma = 0$, some discriminatory information leaks: in Fig.~\ref{fig:KM_conv2d_g0}, for example, we see that the 1st centroid matches the 2nd discriminatory centroid, referring in this case to the asian population.

\subsection{Average mutual information between the discriminatory features}
In order to deepen our analysis, let us try to compute an average information flowing from the NDRd features and let us verify whether the effect of NDR really hides the information flowing from features belonging to the same discriminatory class. In order to move down such a path, let us use a similar approach as in~\cite{tartaglione2020pruning}: in ReLU-activated networks we can associate two different states to the output of a neuron: \textit{on} when the output $y_{l,i}>0$, \textit{off} otherwise. According to this, it is possible to compute the average mutual information between any two ReLU-outputs $\mathbf{y}_1$ and $\mathbf{y}_2$:
\begin{align}
    I(\mathbf{y}_1, \mathbf{y}_2) = \frac{1}{N} \sum_i & \left\{ \Theta(y_{1,i}) \Theta(y_{2,i}) +\right. \nonumber\\
    &\left . +\left[1-\Theta(y_{1,i})\right] \left[1-\Theta(y_{2,i}) \right]\right\}
    \label{eq:I}
\end{align}
where $\Theta(\cdot)$ is the one-step function. Hence, it is possible to average \eqref{eq:I} for the examples belonging between the same class or different classes. As it is possible to observe, the definition of \eqref{eq:I} is very similar to \eqref{eq::regucond}: indeed, NDR is a differentiable proxy of \eqref{eq:I}, which is a non-differentiable quantity.\\
Computing the average mutual information between discriminatory classes is a computational-expensive operation, and we are going to perform it on the MNIST case, with $\Pi_2=\texttt{fc2}$, as described in Sec.~\ref{sec:MNIST} and presented in Fig.~\ref{fig:KM_fc1}.\\
\begin{figure}
    \begin{center}
        \centering
        \includegraphics[width=\columnwidth]{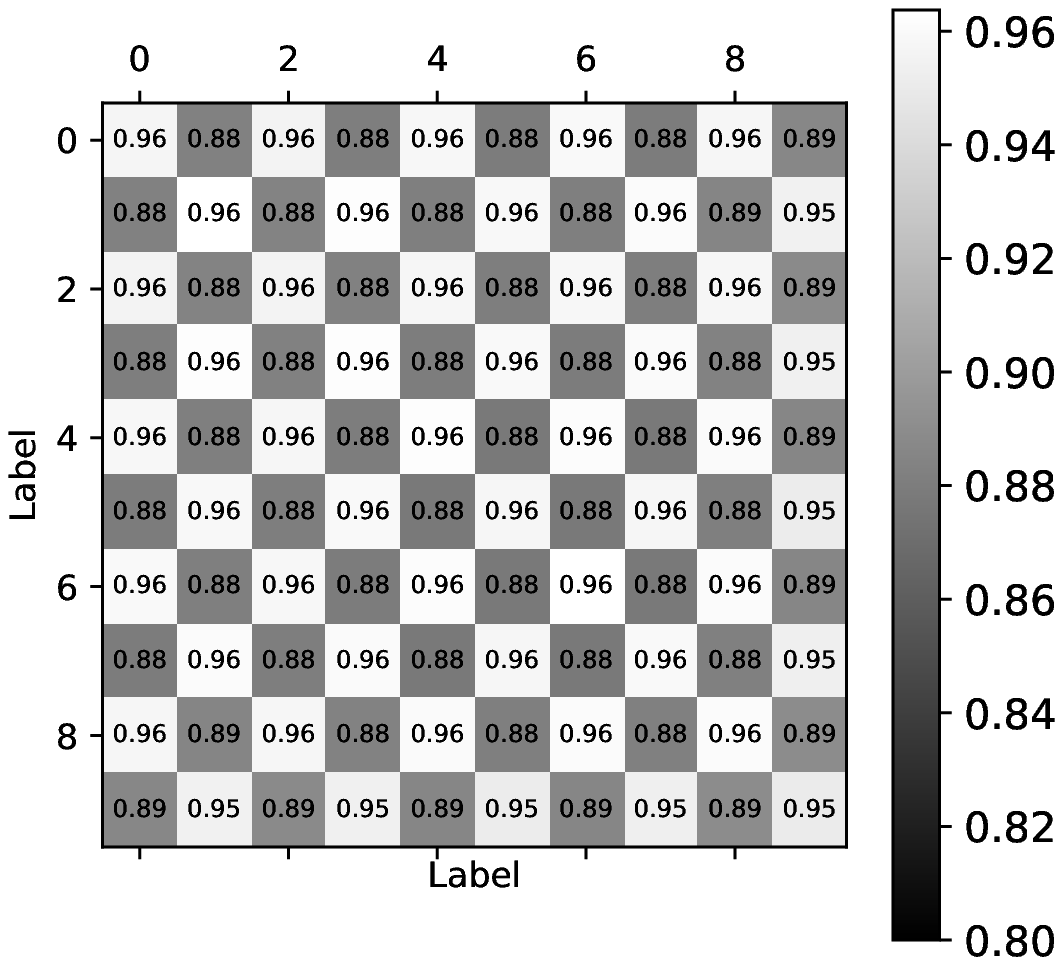}
        \caption{Normalized averaged mutual information between NDRd patterns in the MNIST case.}
        \label{fig:H_fc1}
    \end{center}
\end{figure}
\noindent Fig.~\ref{fig:H_fc1} shows the average mutual information between the 10 discriminatory classes in the MNIST dataset. The purpose of NDR is to make impossible to retrieve the information of the discriminatory class. Such a behavior is observed when the mutual information between different classes is similar to the one the class has itself (the diagonal elements in Fig.~\ref{fig:H_fc1}). Interestingly, we observe a checkerboard-like distribution of the class' mutual information. In particular, there is a strong mutual information within ``odd'' and ``even'' numbers groups, which is the final classification goal of the entire model: however, there is a very small, or even no variance between the mutual information of the discriminatory classes within the same group, making impossible to recover the original discriminatory information. From the figure, it is evident that we have two clusters, represented by the two groups odd/even. This is the effect of NDR, which makes the samples of the same discriminatory class as orthogonal as possible, and then, following the loss minimization, the feature groups, completely non-discriminatory according to the definition of ``discriminatory class'', are naturally formed. Such a behavior has two benefits: the discriminatory class is buried in the classification group (or between classification groups, like in UTKFace) and the information for the main learning task still flows.
\section{Conclusion}
\label{sec:conclusion}
In this work we have proposed NDR, a non-discriminatory regularization term, aiming to hide some defined ``discriminatory'' information during the learning process. In particular, a part of the model learn to filter the information which is seen by the rest of the network. In this way, the ANN model is forced to select other features than the discriminatory ones (like, for example, the ethnicity) to learn a given target task.\\
This work aims at giving a first contribution towards trustworthy AI, providing NDR, a training tool to guarantee, for example, trained models that are not discriminatory against ethnicities. Of course, this comes at a cost: averagely, we observe a drop in the performance. This is understandable, as we are hiding some information potentially useful to solve the task, and it is a trade-off we are aware of.
NDR succeeds in hiding discriminatory features using a very local information (the one contained in the single minibatch), as also observed measuring the mutual information between discriminatory classes. Future work include NDR extension to momentum-based techniques and the formulation of an automatic threshold to find an optimal trade-off able to guarantee non-discrimination while maximizing the performance of the trained model.

\bibliographystyle{IEEEtran}
\bibliography{root}

\end{document}